\pdfoutput=1

\documentclass[11pt]{article}

\usepackage[final]{acl}

\usepackage{times}
\usepackage{latexsym}
\usepackage{subcaption}
\usepackage{longtable}
\usepackage{booktabs} 
\usepackage{amsmath} 
\usepackage{multirow} 
\usepackage{enumitem}
\usepackage{array}
\usepackage{tipa}

\usepackage[T1]{fontenc}

\usepackage[utf8]{inputenc}

\usepackage{microtype}

\usepackage{inconsolata}

\usepackage{graphicx}

\title{Exploring Automated Keyword Mnemonics Generation \\ with Large Language Models via Overgenerate-and-Rank}

\author{
\bf Jaewook Lee, Hunter McNichols, Andrew Lan \\
  University of Massachusetts Amherst \\
  \texttt{\{jaewooklee,wmcnichols,andrewlan\}@cs.umass.edu} \\
  }

\begin{document}
\maketitle
\begin{abstract}
In this paper, we study an under-explored area of language and vocabulary learning: keyword mnemonics, a technique for memorizing vocabulary through memorable associations with a target word via a verbal cue. Typically, creating verbal cues requires extensive human effort and is quite time-consuming, necessitating an automated method that is more scalable. We propose a novel overgenerate-and-rank method via prompting large language models (LLMs) to generate verbal cues and then ranking them according to psycholinguistic measures and takeaways from a pilot user study. To assess cue quality, we conduct both an automated evaluation of imageability and coherence, as well as a human evaluation involving English teachers and learners. Results show that LLM-generated mnemonics are comparable to human-generated ones in terms of imageability, coherence, and perceived usefulness, but there remains plenty of room for improvement due to the diversity in background and preference among language learners.
\end{abstract}

\section{Introduction}
Recent advances in natural language processing have expanded the exploration of developing automated methods for applications in \emph{language learning}, including second language acquisition~\cite{zhang2021negation,yeung2021character,okano2023generating}, language assessment and correction~\cite{katinskaia2021assessing}, and language practice through conversational chatbots~\cite{tyen2022towards,liang2023chatback}. In this work, we investigate an intriguing but relatively under-explored vocabulary learning application: keyword mnemonics (KM)~\cite{atkinson1975application}.

KM is a widely utilized technique employed by language learners to memorize vocabulary effectively, through the creation of memorable associations with keywords that they already know, aided by verbal and visual cues. This technique is used in many language learning resources, from traditional books~\cite{burchers2000vocabulary,heisig2011remembering,Geer2018picturethese} to modern community-based learning platforms~\cite{mnemonicdictionary,koohii}, where users contribute mnemonics and the community vote on their effectiveness.

\begin{figure}[t]
\centering
\includegraphics[width=1\linewidth]{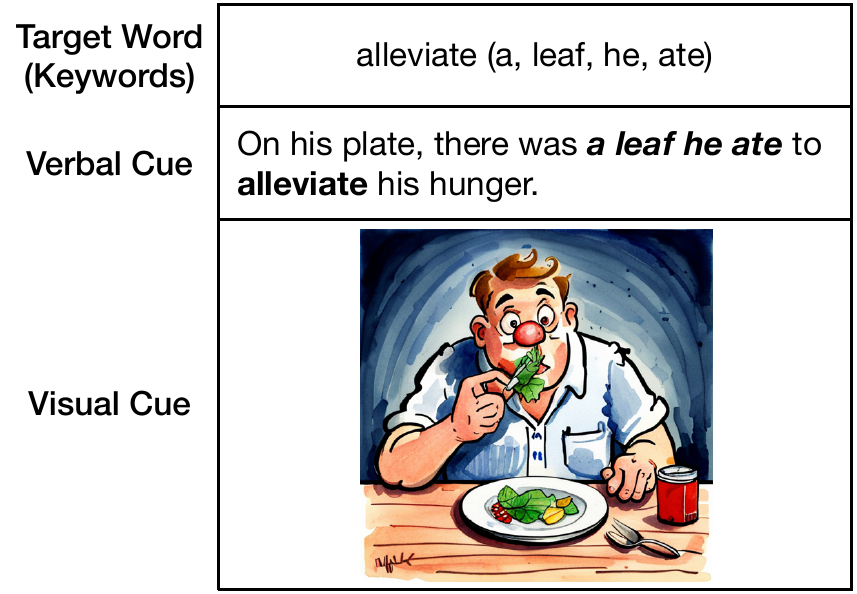}
\caption{An example of human-authored verbal cue~\cite{Geer2018picturethese} for the target word \textit{``alleviate''} and visual cue generated by Stable Diffusion XL. We study automated verbal cue generation in this work and leave visual cue generation for future work.} \vspace{-.5cm}
\label{fig:alleviate_example}
\end{figure}

KM involves a two-step process that creates an acoustic and imagery link for the learner. For instance, consider a learner learning the target word \textit{``alleviate''}, as shown in Figure~\ref{fig:alleviate_example}. First, an acoustic link is established to an already-known set of keywords with similar pronunciation, e.g., \textit{``a, leaf, he, ate.''} Second, a imagery link is established to these keywords with a verbal cue that evokes a mental image, e.g., \textit{``On his plate, there was a leaf to alleviate his hunger.''} These links can be further reinforced using a visual cue.

Despite its effectiveness, KM generation is challenging since it demands significant human effort to meticulously find keywords that resemble the target word and create memorable verbal cues. For instance, the word \textit{``alleviate''} has a large number of possible keyword combinations, such as \textit{``a, leaf, he, ate''} or \textit{``a, levy, it''}; sorting through them takes considerable effort from teachers and learners. Moreover, the task of creating verbal cues using the keywords that evoke vivid mental images adds an extra layer of complexity to the process.


Existing work on automated KM generation remains limited in both technique and evaluation. For example, earlier methods are restricted to generating a single keyword in the context of second language acquisition. The work in~\citet{savva2014transphoner} uses a cognitive psychology-inspired method to automatically generate a keyword in the first language that has high phonetic, orthographic, and semantic similarity with the target word in the second language. 
More recently, the emergence of large language models (LLMs) provides a new and potentially more scalable solution to automated KM generation~\cite{lee2023smartphone}, expanding automated KM generation to generating verbal cues using the keywords generated by \citet{savva2014transphoner}.


However, evaluating KMs remains challenging due to two main reasons: the lack of automated metrics to evaluate the quality of verbal cues, and the subjective nature of mnemonics to language learners. We need automated ways to evaluate important aspects of KMs such as imageability and coherence. The work in~\citet{wu2023composition} explored using text-to-image models for the automatic assessment of sentence imageability, although it has not been investigated in the context of language learning. 

\paragraph{Contributions}
In this paper, we propose an overgenerate-and-rank method to generate verbal cues for vocabulary learning in first language acquisition (English) using LLMs. For a target word that a learner needs to learn, our method first prompts an LLM to generate a set of syllabic keywords and then generate a corresponding verbal cue. For both keywords and verbal cues, we overgenerate and then rank the candidates, using various ranking criteria that are grounded in both cognitive psychology principles used in previous studies and additional insights gained from a pilot user study. 
%
To assess the quality of verbal cues generated by our method, we conduct both 1) an automated evaluation of their imageability and coherence using proxy metrics and 2) a human evaluation with both English teachers and learners, comparing LLM-generated and human-authored cues on imageability, coherence, and additionally usefulness to language learners. 
Results indicate that our LLM-generated cues are comparable to (and often better than) human-authored ones. We also conduct several case studies and discuss the varying degrees of human agreement on different aspects of KMs, which highlight the need to further align generated KMs with individual preferences. 

\section{Problem Statement}
We now formally define the verbal cue generation task. To do so, we first need to generate a set of \emph{syllabic keywords} given a target word, then a \emph{verbal cue} that contains them. 
Given a target word $t$, we denote its syllables as $\mathcal{S}_t = (s_1, ..., s_L)$ where $L$ is the total number of syllables in $t$. Our goal is to generate a set of $M$ syllabic keywords as $\mathcal{K} = \{k_1, \ldots, k_M\}$, where $k_m$ is phonetically similar to one (or more) consecutive syllables in $t$, beginning at index $l_m$ and ending at $l'_m$, which we denote as $s_{l_m:l'_m}$. The set of syllabic keywords also has to cover all syllables, i.e., $\cup_m \{l_m, \ldots, l'_m\} = \{1, \ldots, L\}$. This task is challenging, since that there are $2^{L-1}$ total possible ways to split the set of syllables of $t$, and that we need to choose the right keywords that preserve the phonetic properties of the target word. 
Then, our task is to generate a verbal cue $\mathcal{V} = (w_1,...,w_N)$, where $w_i$ denotes the $i$-th word in the cue. 
The constraint we put on the verbal cue is that it must contain both the specific target word and all syllabic keywords, which we formally define as
\begin{equation*}
\forall x \in \mathcal{K} \cup \{t\}, \, \exists i \in \{1,\ldots,N\} \, \text{s.t. } w_i = x.
\end{equation*}

\section{Methodology}
We propose a two-step overgenerate-and-rank method for KM generation via LLM prompting to navigate the large space of possible keywords and verbal cues. First, we overgenerate multiple sets of candidate syllabic keywords by prompting an LLM, rank them according to a series of measures, and select the top set. Second, we use these keywords to overgenerate multiple verbal cues, rank them according to insights obtained from a pilot study with English teachers. See Supplementary Material~\ref{sec:prompt} for the exact prompts we used. 

\subsection{Keyword Generation}
\label{sec:keyword_gen}
For keyword generation, we craft a prompt that includes the task description, instructing the LLM to generate a set of keywords phonetically similar to the syllables of the target word. The instructions include the following rules: 1) Each keyword must be a complete word, familiar and commonly used at an SAT vocabulary level; 2) The keywords should resemble the target word phonetically when spoken together, even if they don't match the exact number of syllables; 3) The words must not be offensive. We also provide a set of three in-context examples. 

We overgenerate up to $2L+1$ sets of candidate keywords $\hat{\mathcal{K}}$ where $L$ is the number of syllables in the target word. To account for aspects of keyword generation other than phonetic similarity, we also consider the following psychology measures used in prior work~\cite{savva2014transphoner} when ranking keywords: imageability, orthographic similarity, which measures how closely the keywords resemble the spelling of the target word, and semantic similarity, which measures how closely are the meanings of the keywords to that of the target word.

To create the imageability ranking $R_{\text{img}}$, we compute the average imageability score~\cite{ljubesic18predicting,scott2019glasgow} of the keywords, excluding stopwords. The score $f_\text{img}$ is determined by the sum of imageability scores ($\mathcal{L}_\text{img}$) of individual keywords divided by the number of keywords:
\begin{equation*}
    f_\text{img}(\mathcal{\hat{K}}) = 1/|\mathcal{\hat{K}}| \textstyle \sum_{k \in \mathcal{K}}\mathcal{L}_\text{img}[k].
\end{equation*}
A set of keywords is ranked higher in $R_{\text{img}}$ if the imageability score is higher; same for the other rankings below unless stated otherwise. 
To create the orthographic similarity ranking $R_{\text{orth}}$, we calculate the Levenshtein distance  between a target word and the concatenated keywords, $D_\text{lev}(\cdot,\cdot)$. The score $f_\text{orth}$ is determined by this distance:
\begin{equation*}
    f_\text{orth}(\mathcal{\hat{K}}, t) = D_\text{lev}(\text{concat}(\mathcal{\hat{K}}), t).
\end{equation*}
To create the semantic similarity ranking $R_{\text{sem}}$, we compute the cosine similarity between each keyword's embeddings and the target word's embedding~\cite{bojanowski2017enriching}. The score $f_\text{sem}$ is determined by this maximum similarity:
\begin{equation*}
    f_\text{sem}(\mathcal{\hat{K}}, t) = \underset{k \in \mathcal{\hat{K}}}{\max} \, \cos(\text{emb}(k), \text{emb}(t)).
\end{equation*}


The final, overall keyword ranking is defined as the geometric mean of the three rankings, i.e., $\sqrt[3]{\left(R_{\text{img}} \cdot R_{\text{orth}}\cdot R_{\text{sem}}\right)}$. We observe empirically that this method works well and combines the rankings without additional parameters to tune. Also, ties in this ranking are rare and we break randomly.


\subsection{Pilot Study for Verbal Cue Generation}
\label{sec:pilot_study}
Since the perceived usefulness of a verbal cue can be highly subjective to each learner and not extensively studied in prior work, we conduct a pilot study with English teachers on both LLM-generated and human-authored verbal cues, to identify important features of verbal cues that they think would help learners. In this pilot test, we use a single in-context example, prompting an LLM for both keywords and a corresponding verbal cue at once. We received four main suggestions that inform our verbal cue generation process:

\begin{enumerate}[label=\roman*., itemsep=-0.5ex]
    \item Keep the original keyword order in the cue.
    \item Provide clear context for the target word.
    \item Use words at the same (complexity) level as or lower than the target word.
    \item Keep the cue short; long ones are not helpful.
\end{enumerate}

\subsection{Verbal Cue Generation}

For verbal cue generation, we craft a prompt with a task description that instructs the LLM to generate a coherent verbal cue containing the target word and all syllabic keywords. In each in-context example, we also include the target word's meaning to avoid homonyms, together with a context explanation for richer contextual information. 



We overgenerate up to 5 candidate verbal cues and filter out cases where the syllabic keywords are not in order, according to suggestion i). To account for other suggestions, we rank these candidate verbal cues according to two measures: context completeness and age of acquisition (AoA). The former refers to the extent of contextual information on the target word given by the verbal cue, while the other is a psycholinguistic measure that indicates the typical age at which a word is learned. 

To create the context completeness ranking $R_{\text{cont}}$, we employ a masked modeling technique inspired by suggestion ii). Specifically, we mask out the target word $t$ within a verbal cue $\mathcal{\hat{V}}$ and prompt an LLM to predict the five most likely words under the mask, which are listed in a set $\mathcal{C}$:
\begin{equation*}
    \mathcal{C} = \text{LLM}_\text{top-5}\left(\text{mask}(\mathcal{\hat{V}}, t)\rightarrow t\right).
\end{equation*}
Then, we calculate the average cosine similarity between the word embeddings of each predicted word $c$ and the target word $t$:
\begin{equation*}
    f_\text{cont}(\mathcal{C}, t) = 1/|\mathcal{C}| \textstyle\sum_{c \in \mathcal{C}} \cos \left(\text{emb}(c), \text{emb}(t)\right).
\end{equation*}
Intuitively, a high average cosine similarity means that it is easy for the LLM to predict the target word (or similar words) given other words in the verbal cue, which indicates that it contains complete contextual information. 
To create the AoA ranking $R_{\text{AoA}}$, we sum the AoA ($\mathcal{L}_\text{AoA}$)~\cite{kuperman2012age} of words in the verbal cue to establish a ranking, which penalizes complex words in the verbal cue according to suggestion iii). The sum also penalizes long verbal cues according to suggestion iv). The word complexity score $f_\text{AoA}$ is given by:
\begin{equation*}
    f_\text{AoA}(\hat{\mathcal{V}}) =  \textstyle \sum_{w \in \mathcal{\hat{V}}}\mathcal{L}_\text{AoA}[w]. 
\end{equation*}
Unlike all other rankings above, verbal cues with lower AoA scores are ranked higher in $R_{\text{AoA}}$. 

The final, overall verbal cue ranking is defined as the geometric mean of the two rankings, i.e., $\sqrt{\left(R_{\text{cont}} \cdot R_{\text{AoA}}\right)}$.

\section{Automated Evaluation}
In the following sections, we aim to explore three specific aspects of KMs that can vary between individuals: \textit{imageability}, \textit{coherence}, and \textit{usefulness}. In this section, we introduce proxy metrics for automatically evaluating imageability and coherence. 
Later, in Section~\ref{sec:human_eval}, we conduct a human evaluation to additionally measure the perceived usefulness of verbal cues by both English teachers and learners, in addition to imageability and coherence. 
%




\subsection{Dataset}
Since there is no established baseline for evaluating the three aspects of KMs, we compare with human-authored cues with LLM-generated ones. We utilize the book ``Picture These SAT Words!''~\cite{Geer2018picturethese}, which consists of around $300$ SAT words and provides keywords, verbal, and visual cues. We randomly sample $60$ target words from this book to use in our experiment. \emph{We emphasize that due to the high cost of human evaluation, 60 words is the maximum scale (1/5 of words in the book) that we can conduct this initial study at.} 

We use GPT-4 (\texttt{temp}$=0.7$, \texttt{top\_p}$=1$) to generate both keywords and verbal cues via overgenerate-and-rank. See Supplementary Material~\ref{sec:verbal_cues} for all target words used in our evaluation along with LLM-generated and human-authored verbal cues. 

\subsection{Metrics}
We define the evaluation criteria for the three aspects of verbal cues as follows: Imageability assesses the effectiveness of verbal cues in evoking mental images, coherence evaluates the logical consistency of the verbal cues, and usefulness determines how helpful these cues are in aiding a learner to learn the target word. 

We employ several automated metrics as proxies for assessing the imageability and coherence of verbal cues. Alongside these metrics, we also check the quality of keywords to compare those generated by LLM with human-authored ones, thereby performing a preliminary validation.


\subsubsection{Keywords}

\begin{table*}[ht]
\centering
\scalebox{.85}{
\begin{tabular}{c|cc|cccc}
\toprule
Method  & $\text{Syllable Ratio}^\uparrow$ & $\text{Phonetic Sim.}^\uparrow$ & $\text{Imageability}^\uparrow$  & $\text{Orthographic Sim.}^\uparrow$ & $\text{Semantic Sim.}^\uparrow$ \\ 
\midrule
Barron        & 0.87 & \textbf{0.52}            & 0.51           & 0.37          & 0.11   \\
\textbf{Ours}  & \textbf{0.92} & \textbf{0.52}  & \textbf{0.76}  & \textbf{0.40} & \textbf{0.12}  \\
\bottomrule
\end{tabular}
}
\caption{Comparative analysis of syllabic keywords from our method and those in Barron's book \cite{Geer2018picturethese}, using the three ranking criteria along with two additional metrics: syllable ratio and phonetic similarity. Values are normalized to $[0,1]$ for ease of comparison.}
\label{tab:key_comp}
\end{table*}
We evaluate the quality of keywords by applying three ranking criteria outlined in Section~\ref{sec:keyword_gen}: \textbf{Imageability} (word-level), orthographic similarity (\textbf{Orthographic Sim.}), and semantic similarity (\textbf{Semantic Sim.}), adopted by prior work~\cite{savva2014transphoner}. We also introduce two new criteria: \textbf{Syllable Ratio} and phonetic similarity (\textbf{Phonetic Sim.}). Syllable ratio is calculated by dividing the number of keywords by the total number of syllables, assessing how well the keywords align with the syllables of the target word. Phonetic similarity, determined using the International Phonetic Alphabet (IPA)~\cite{international1999handbook}, calculates the Levenshtein distance between the concatenated IPAs of the keywords and that of the target word. 

\subsubsection{Verbal Cue}
For imageability, we adopt a similar methodology to the one introduced in~\cite{wu2023composition}, which generates images from textual sentences using a text-to-image model DALL$\cdot$E mini~\cite{dayma2021dall} and applies the text-image alignment metric CLIP~\cite{radford2021learning}. In our study, we use a larger text-to-image model, Stable Diffusion 2.0 ~\cite{rombach2021highresolution}, along with a more advanced text-image alignment metric, ImageReward~\cite{xu2024imagereward}. Given a verbal cue \(\hat{\mathcal{V}}\), we randomly sample a set of $9$ images $\mathcal{I}$ using Stable Diffusion. The imageability of the verbal cue is then set as the maximum ImageReward score (\textbf{IMR}) between the verbal cue and the generated images. For coherence, we calculate the perplexity (\textbf{PPL}) of the verbal cue using the open-source LLM Llama3-8B~\cite{touvron2023llama}, since textual coherence correlates with how well a pre-trained LLM can predict a sequence of text tokens.

\begin{table}[t]
\centering
\scalebox{.85}{
\begin{tabular}{cc|cccc}
\toprule
\multicolumn{2}{c|}{Method}  & \multirow{2}{*}{$\text{IMR}^\uparrow$} & \multirow{2}{*}{$\text{PPL}^\downarrow$ }\\ 
\cmidrule(lr){1-2}
             Keyword     & Verbal Cue                     &         &         \\ 
\midrule

\multicolumn{2}{c|}{Barron}                         & 0.56    & 444.8   \\
\multicolumn{2}{c|}{\textbf{Ours}}                        & \textbf{0.61}    & \textbf{156.4}   \\ 
\bottomrule
\end{tabular}
}
\caption{Comparative analysis of verbal cues from our method and Barron's book \cite{Geer2018picturethese}, using the ImageReward and Perplexity metrics.}
\label{tab:verbal_comp}
\end{table}

\subsection{Results}
In what follows, ``Ours'' refers to LLM-generated keywords and verbal cues using our overgenerate-and-rank method, whereas ``Barron'' refers to human-authored cues in Barron's book.

\subsubsection{Keywords}
Table 1 compares keyword generation performance between our method and humans across all metrics. We see that our keywords are as good or better than human-authored ones across all metrics. In terms of imageability, our method gets a much higher score than human authors, likely because human authors often employs proper nouns (e.g., \textit{``Guy''} for \textit{``beguile''}) or alphabet-based terms (e.g., \textit{``D grade''} for \textit{``degradation''}) that are less imageable. On the other metrics, our method slightly outperforms human authors, with the exception of matching human authors on phonetic similarity. 
As a case study, for the keyword \textit{``enmity''} (IPA: /\textipa{'\textepsilon nmIti}/), the LLM-generated \textit{``hen, mitt, tee''} (IPA: /\textipa{hEn}/, /\textipa{mIt}/, /\textipa{ti}/) and the human-authored \textit{``N, mitt, hi''} (IPA: /\textipa{\textepsilon n}/, /\textipa{mIt}/, /\textipa{hi}/) produces keywords with the same Levenshtein distances in the IPA space from the target word. This example suggests that human-authored ones often choose less imageable words, (e.g., \textit{``N''}), to meet strict phonetic similarity criteria, whereas LLMs, which are not specifically trained on phonetic data, tend to select more common and imageable words. Therefore, using LLM-generated keywords may reduce human effort on finding keywords for verbal cues that are not only phonetically similar but also imageable.

\subsubsection{Verbal Cue} 
Table~\ref{tab:verbal_comp} compares verbal cues generated by our overgenerate-and-rank method against human-authored cues in Barron's book. We see that our method outperforms human authors on both imageability and coherence, especially on the latter, where human-authored verbal cues in Barron's book have much higher perplexity than LLM-generated ones. This result suggests that our method produces verbal cues that are not only vivid but also more coherent compared to Barron's book. 
One possible reason for these results is that human-authored cues often make up unrealistic scenarios, thus significantly decreasing their coherence. 
%
For example, the highest perplexity score we observe is for the human-authored cue \textit{``A polemical polar Mick call''} for the word \textit{``polemical,''} scoring $3303.9$. This cue evokes the image of someone in the Arctic, angrily using a phone. 
The phrase \textit{``polemical polar Mick''} is a highly unusual combination; in contrast, the LLM-generated cue, \textit{``The polemical John, standing like a pole, accuses me of having ideas as dark as coal,''} creates a more natural scenario with a more coherent sentence, resulting in a much lower perplexity score of $88.6$. We provide a more detailed, per-word analysis in Section~\ref{sec:cs} on LLM-generated cues and also discuss feedback from real English learners.

\subsection{Ablation}

We perform an ablation study to assess the impact of several aspects of our verbal cue generation method. First, we assess the effectiveness of a two-stage pipeline, i.e., not generating keywords and using human-authored ones for verbal cue generation instead. Second, we evaluate the performance of a smaller, open-source language model, Llama 3-8B~\cite{touvron2023llama}. For this model, we conduct two experiments: fine-tuning with Barron's book (\textbf{FT}), and prompting with the same prompt as our method (\textbf{Ours}). See Supplementary Material~\ref{sec:sp_ablation} for the prompt and detailed model configuration.

\begin{table}[t]
\centering
\scalebox{.85}{
\begin{tabular}{cc|cccc}
\toprule
\multicolumn{2}{c|}{Method}  & \multirow{2}{*}{$\text{IMR}^\uparrow$} & \multirow{2}{*}{$\text{PPL}^\downarrow$ }\\ 
\cmidrule(lr){1-2}
             Keyword     & Verbal Cue                       &         &         \\ 
\midrule
\multirow{2}{*}{Barron}  & $\text{Llama3}_{\text{FT}}$      & 0.53    & 482.0       \\ %
                         & $\text{GPT-4}_\text{Ours}$       & 0.54    & 147.1     \\  
\multirow{2}{*}{Ours}    & $\text{Llama3}_{\text{Ours}}$    & 0.58    & 257.7       \\
                         & $\text{GPT-4}_\text{Ours}$       & 0.61    & 156.4     \\   
\bottomrule
\end{tabular}
}
\caption{Ablation study changing our two-stage pipeline and the underlying LLM, evaluated on ImageReward and Perplexity. $\text{GPT-4}_\text{Ours}$ denotes our method.}
\label{tab:ablation}
\end{table}

Table~\ref{tab:ablation} shows the ablation study results. We see that using our method to generate sets of keywords leads to much higher imageability in downstream verbal cue generation, compared to using human-authored keywords from Barron's book. We note that human-authored keywords do result in slightly better coherence, indicating a trade-off, however the significantly better imageability tilts the balance towards LLM-generated keywords. Meanwhile, we also see that Llama 3 results in lower performance on both metrics compared to GPT-4, even after fine-tuning on human-authored verbal cues. Notably, $\text{Llama3}_{\text{Ours}}$ successfully followed the prompt instructions only 58\% of the time (either missing keywords or the target word), whereas $\text{Llama3}_{\text{FT}}$ followed them 90\% of the time. This result suggests that automated KM generation is perhaps a task that is too difficult for smaller LMs to do well on, especially without fine-tuning on real KMs.

\section{Human Evaluation}
\label{sec:human_eval}
\subsection{Setup}
To further evaluate the imageability and coherence of KMs, and more importantly, their usefulness to language learners, we conduct human evaluation from both teachers' and learners' perspectives. From the teachers' perspective, we employ four evaluators recruited through \citet{upwork}, all of whom have experience teaching English at the high school level or preparing learners for English exams. From the learners' perspective, we hire nine university students through on-campus recruiting, including four freshmen and five sophomores; we require them to have recently took the SAT exams since our target words are SAT-level. 


We conduct a preliminary survey among nine students that ask them to indicate their unfamiliarity with the 60 target words in our evaluation. The word \textit{``polemical''} is identified as the most challenging, with all nine students unfamiliar with it. Similarly challenging words include \textit{``abstemious,''} \textit{``quiescence,''} and \textit{``threadbare.''} On the other end of the spectrum, 32\% of all target words, such as \textit{``aesthetic''} and \textit{``authoritarian''}, are familiar to all students. This result shows the inherent difficulty of our task since, despite having five native English speakers out of nine in our group, some words are challenging even to these college-level individuals, justifying our participant selection criterion.


The experiment was conducted online through a web application. Prior to the experiment, we provided teachers and learners  with a scoring rubric in Section~\ref{sec:sp_human_eval} to calibrate their judgment, given the likely subjectivity among human evaluators on our evaluation criteria, especially usefulness. During the evaluation, the web application showed both LLM-generated and human-authored cues, one at a time. The ordering of target words and the source of the verbal cues (LLM vs.\ human) were randomized and not disclosed to minimize potential biases. 
\subsection{Results}
\begin{table*}[hbt!]
    \centering
    \scalebox{0.8}{
    \begin{tabular}{lcccccccc}
        \toprule
        & \multicolumn{4}{c}{\textbf{5-point Likert Scale}} & \multicolumn{4}{c}{\textbf{Spearman's $\rho$}} \\
        \cmidrule(lr){2-5} \cmidrule(lr){6-9}
        & \multicolumn{2}{c}{Teachers} & \multicolumn{2}{c}{Learners} & \multicolumn{2}{c}{Teachers} & \multicolumn{2}{c}{Learners} \\
        \cmidrule(lr){2-3} \cmidrule(lr){4-5} \cmidrule(lr){6-7} \cmidrule(lr){8-9}
        & Ours & Barron & Ours & Barron & Ours & Barron & Ours & Barron \\
        \midrule
        Imageability & 3.54* (1.25) & 2.77 (1.42) & 3.50* (1.23) & 2.52 (1.25) & 0.24 & 0.50 & 0.15 & 0.32 \\
        Coherence    & 3.60* (1.25) & 2.90 (1.47) & 3.33* (1.25) & 2.58 (1.24) & 0.34 & 0.56 & 0.27 & 0.35 \\
        Usefulness   & 2.89* (1.27) & 2.20 (1.36) & 3.11* (1.34) & 2.29 (1.30) & 0.31 & 0.59 & 0.20 & 0.32 \\
        \bottomrule
    \end{tabular}
    }
    \caption{Comparison of mean (standard deviation) of 5-point Likert scale ratings and average Spearman rank-order correlation on verbal cues across different groups (four teachers and nine learners).}
    \label{tab:comp2}
\end{table*}
Table~\ref{tab:comp2} shows the average of 5-point Likert scale ratings and the Spearman's rank correlation coefficient (Spearman's $\rho$), among teachers and learners, for both LLM-generated and human-authored verbal cues. We see that overall, participants found verbal cues, especially LLM-generated ones, to be imageable and coherent but relatively less useful. A Wilcoxon signed-rank test indicates a statistically significant difference (labeled as * with $p<0.05$ in Table~\ref{tab:comp2}), which shows that LLM-generated cues are preferred over human-authored ones in all cases. 
%
These findings align with those from the automated evaluation (Table~\ref{tab:verbal_comp}), where our overgenerate-and-rank method also score better on both imageability and coherence than human-authored ones. 

From the table, we make two observations: First, teachers show higher inter-rater agreement than learners, as indicated by pearman's $\rho$. 
%
%
This result can be explained by several key differences between the two participant groups. Teachers, with their professional background in education, tend to have a deep understanding of language, real-world pedagogical settings in language learning, and characteristics of many learners they have interacted with. Therefore, their assessments are likely less subjective since they focus on the pedagogical value of KMs across an entire learner population and the broader application of these cues within curricular goals. In contrast, learners, with less language proficiency, are primarily influenced by their personal background and experiences, which can lead to high subjectivity when individual preferences differ. 
%
%
We also find a lower level of agreement, with a much wider range of scores, on LLM-generated cues than on human-authored ones. This result can be explained by the diverse styles among LLM-generated cues spurring a high degree of subjectivity, since LLMs are trained on web-scale textual data, while the book contains verbal cues authored a small group of authors, with similar style.

\section{Case Study}
\label{sec:cs}
We now qualitatively analyze cues generated by our method based on feedback collected from learners through a post-experiment survey. We asked learners to identify and explain their top five most useful and least useful cues. 
From their responses, we discuss four representative examples of the most and least useful verbal cues, as shown in Table~\ref{tab:qual}: the more useful cue being more imageable and coherent (higher IMR, lower PPL), more imageable but less coherent (higher IMR, higher PPL), more coherent but less imageable (lower PPL, lower PPL), and less imageable and coherent (lower IMR, higher PPL). We also show an image $\hat{i}$ used to calculate the IMR score in each case.



\begin{table*}[t!]
\centering
\scalebox{0.85}{
\begin{tabular}{cc}
\scalebox{0.8}{
\begin{tabular}{>{\raggedright\arraybackslash}m{5.5cm} >{\centering\arraybackslash}m{2cm} >{\centering\arraybackslash}m{1cm} >{\centering\arraybackslash}m{1cm}}
\toprule
\multicolumn{4}{c}{\textbf{Most Useful}} \\
\midrule
\multicolumn{1}{>{\centering\arraybackslash}m{5cm}}{Verbal Cue $\hat{\mathcal{V}}$} & Image $\hat{i}$ &  $\text{IMR}^\uparrow$ & $\text{PPL}^\downarrow$  \\
\midrule
In the \textit{art}-loving town, the \textbf{artisan} sips his \textit{tea} under the \textit{sun}. & \includegraphics[width=1.8cm, height=1.8cm]{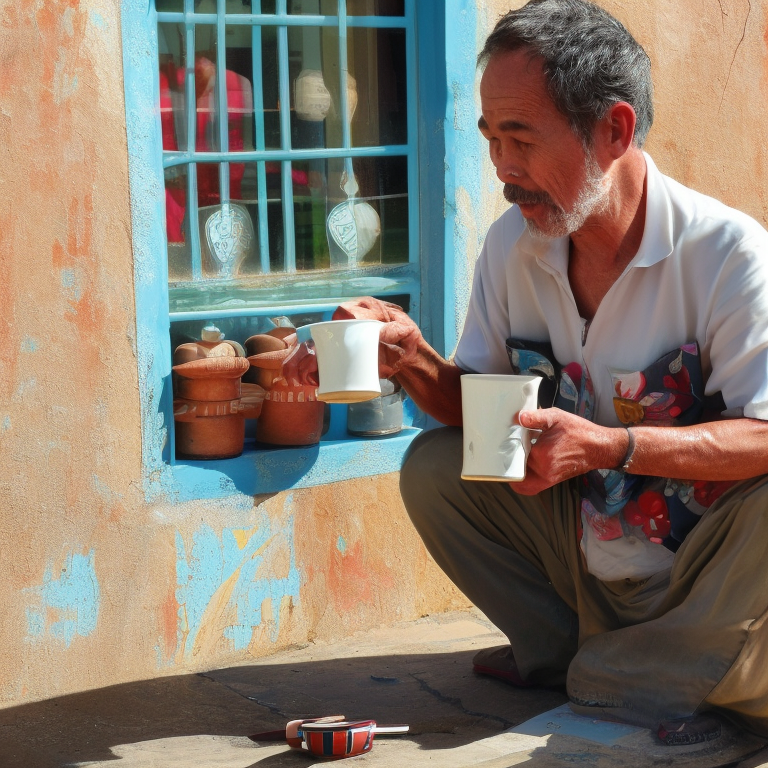} & \textcolor{teal}{1.74} & \textcolor{teal}{86.0} \\
An \textit{ex}-champion \textit{horse} turned \textit{stiff} from his \textbf{exhaustive} training. & \includegraphics[width=1.8cm, height=1.8cm]{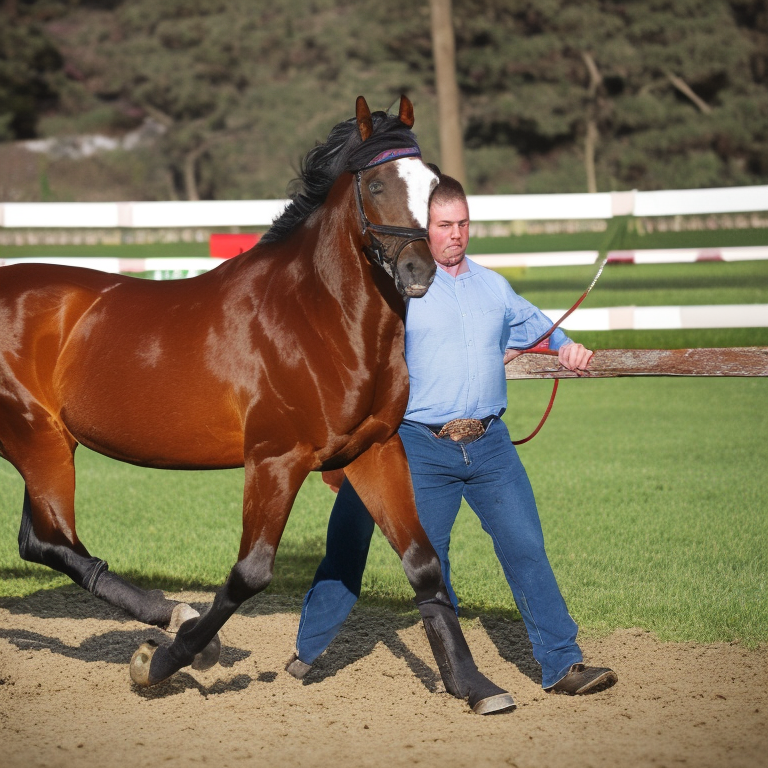} & \textcolor{teal}{1.13} & \textcolor{magenta}{358.7} \\
After \textit{reading} the clean \textit{track} record of the suspect, the detective had to \textbf{retract} his suspicion. & \includegraphics[width=1.8cm, height=1.8cm]{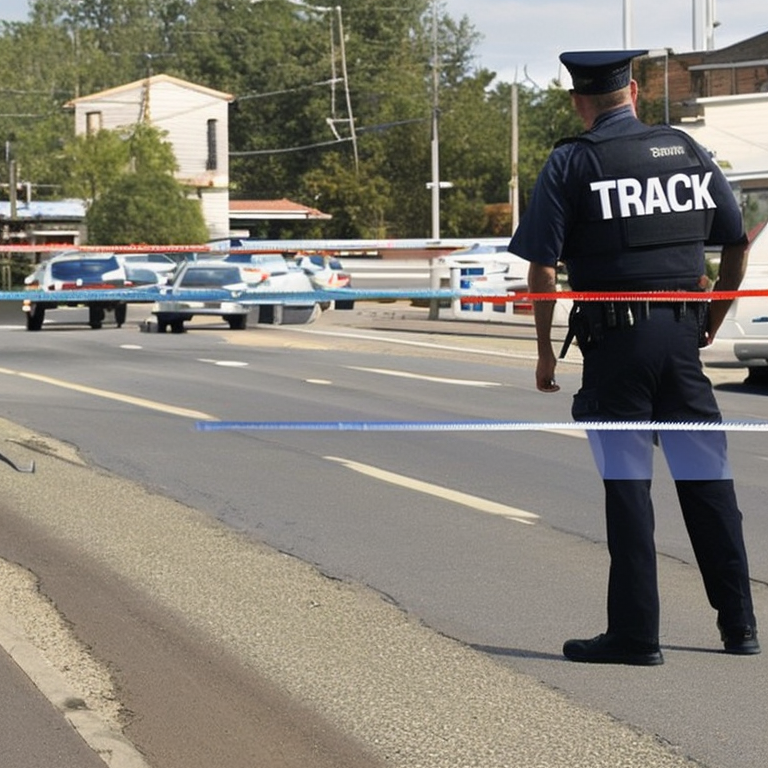} & \textcolor{magenta}{-0.11} & \textcolor{teal}{71.1} \\
Worn to a \textit{thread}, the \textit{bear} became \textbf{threadbare} but still cherished. & \includegraphics[width=1.8cm, height=1.8cm]{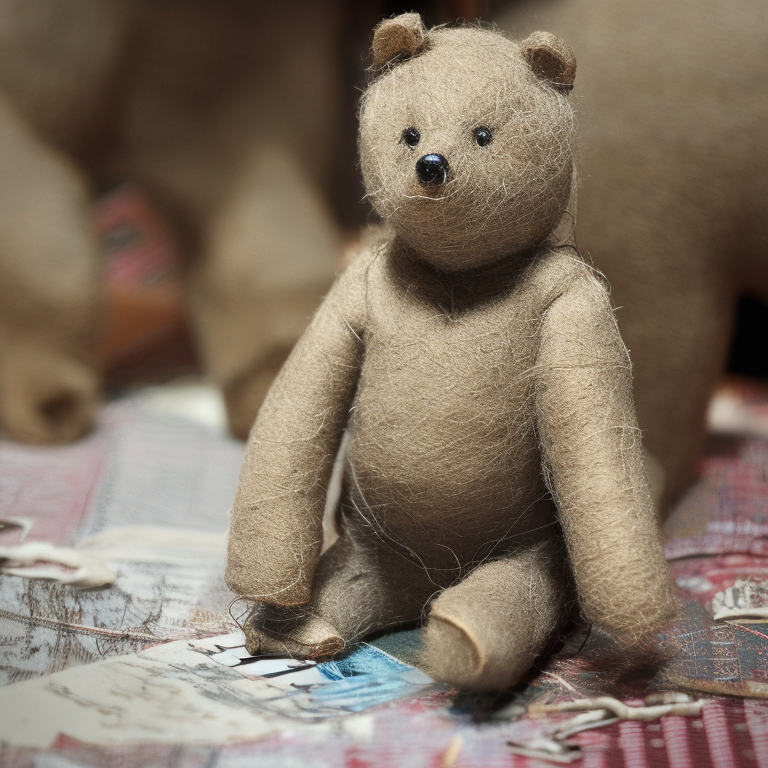} & \textcolor{magenta}{0.44} & \textcolor{magenta}{91.2} \\ 
\bottomrule
\end{tabular}
}
&
\scalebox{0.8}{
\begin{tabular}{>{\raggedright\arraybackslash}m{5.5cm} >{\centering\arraybackslash}m{2cm} >{\centering\arraybackslash}m{1cm} >{\centering\arraybackslash}m{1cm}}
\toprule
\multicolumn{4}{c}{\textbf{Least Useful}} \\
\midrule
\multicolumn{1}{>{\centering\arraybackslash}m{5cm}}{Verbal Cue $\hat{\mathcal{V}}$} & Image $\hat{i}$ & $\text{IMR}^\uparrow$ & $\text{PPL}^\downarrow$ \\
\midrule
A \textit{pair} teased with `what \textit{if} \textit{her} \textit{doll} disappears?', hiding it in \textbf{peripheral} areas, but her attention captured it all. & \includegraphics[width=1.8cm, height=1.8cm]{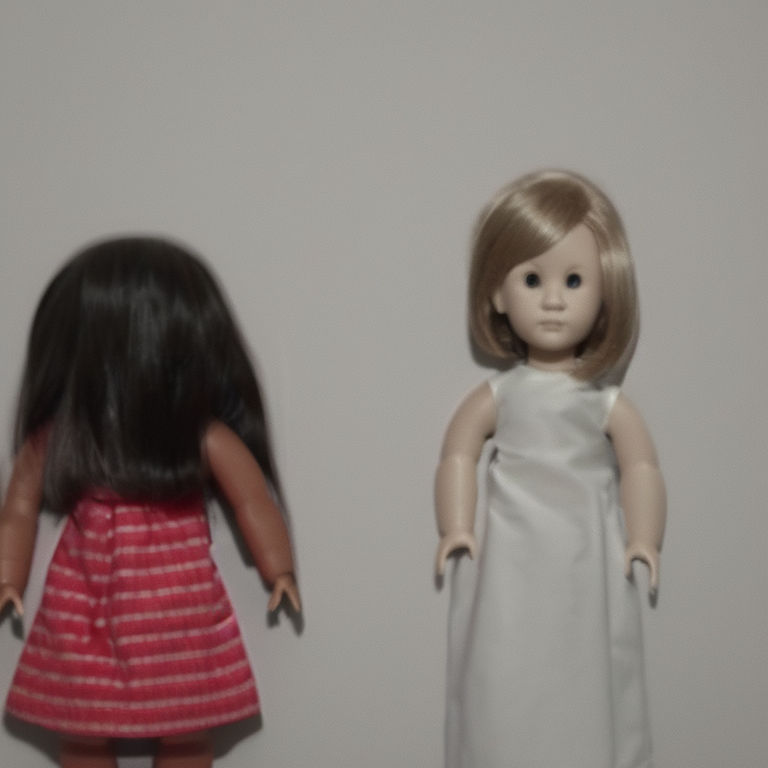} & \textcolor{magenta}{0.08} & \textcolor{magenta}{380.7} \\
Sarah, studying \textbf{phenomena}, was on the \textit{phone} when she had to say, ``\textit{No}, \textit{Ma}.'' & \includegraphics[width=1.8cm, height=1.8cm]{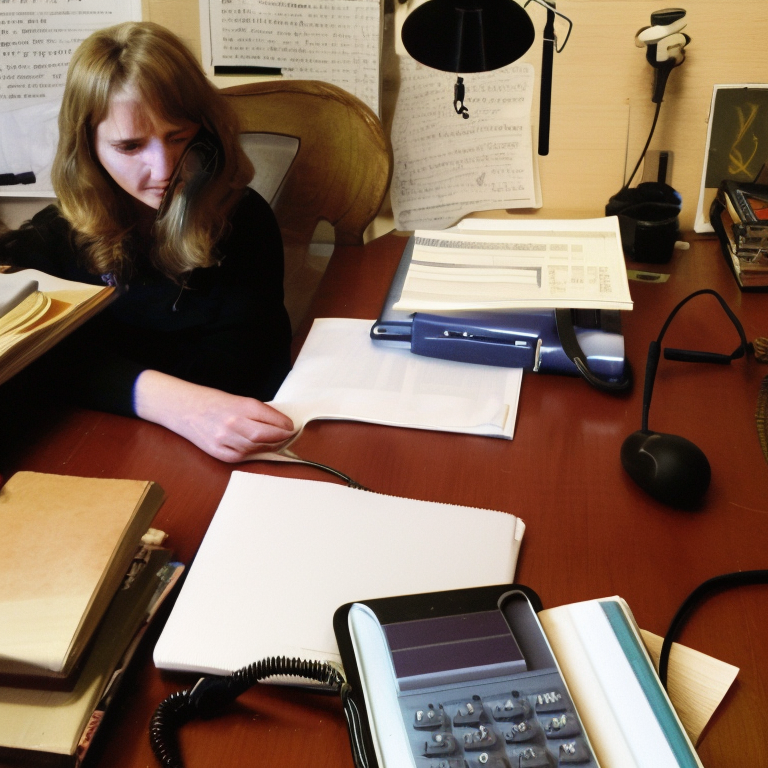} & \textcolor{magenta}{0.07} & \textcolor{teal}{95.3} \\ 
\textit{In} a \textit{tea} gathering, a \textit{mate} walked in to \textbf{intimidate} everyone. & \includegraphics[width=1.8cm, height=1.8cm]{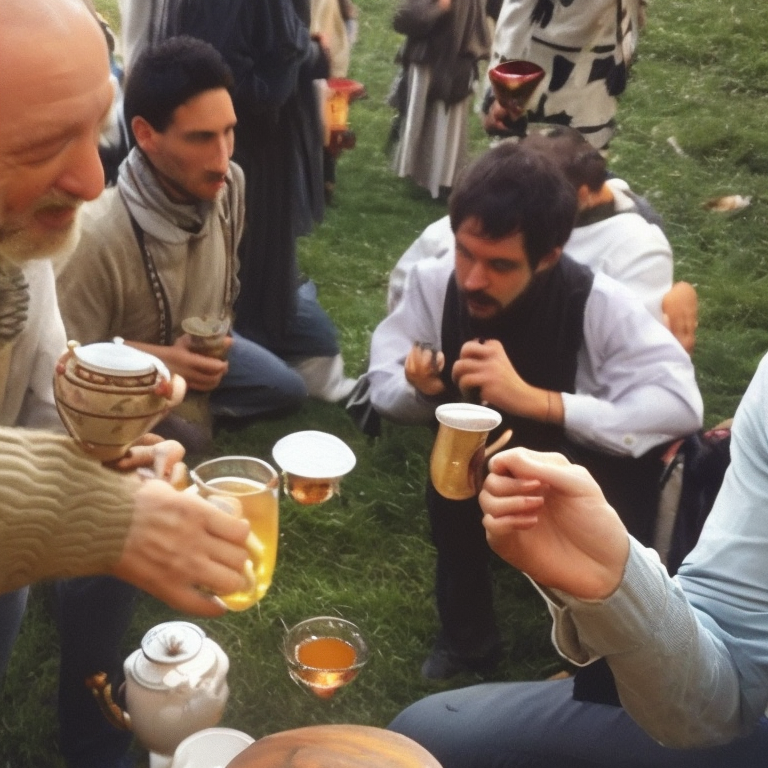} & \textcolor{teal}{0.51} & \textcolor{magenta}{515.3} \\ 
\textit{Veer}'s talent was \textit{too} remarkable, \textit{so} he became a \textbf{virtuoso}. & \includegraphics[width=1.8cm, height=1.8cm]{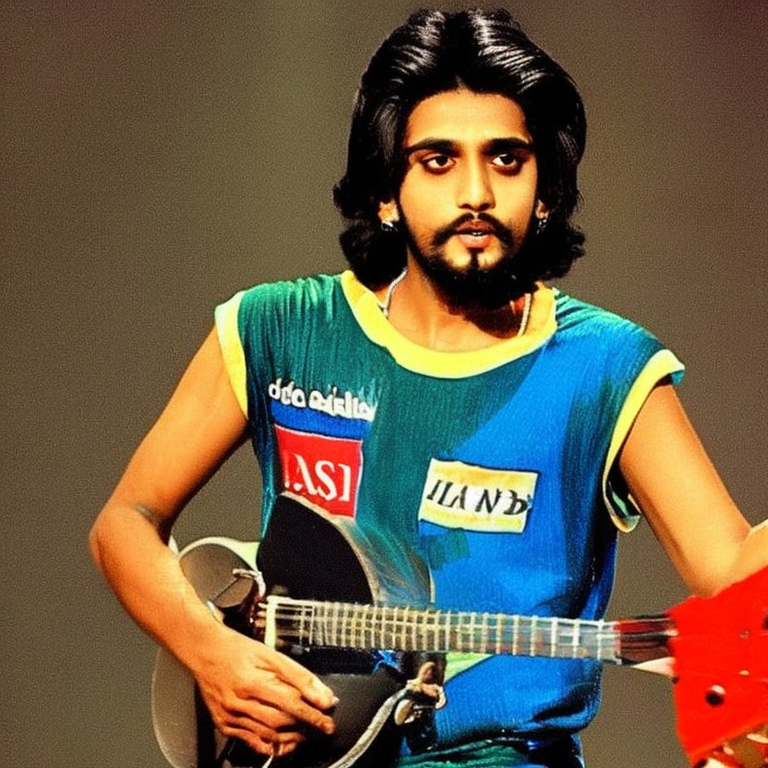} & \textcolor{teal}{1.70} & \textcolor{teal}{80.0} \\
\bottomrule
\end{tabular}
}
\end{tabular}
}
\caption{Most and least useful verbal cues (keywords in \textit{italic} and a target word in \textbf{bold}) indicated by learners. Colors indicate whether a score is higher (\textcolor{teal}{teal}) or lower (\textcolor{magenta}{magenta}).}
\label{tab:qual}
\end{table*}




\noindent\textbf{Highly Imageable \& Coherent}
The keywords for \textit{``artisan''} blend seamlessly into the verbal cue, creating an imageable scene, whereas the keywords for \textit{``peripheral''} struggle to blend into the cue due to their complexity, which arises from the challenge of finding phonetically similar words that also match the syllable count. If this condition is relaxed, the verbal cue can improve: for \textit{``pear, for, all,''} the cue \textit{``A pear tree on the town's peripheral bears fruits for all.''} blends seamlessly in the cue, achieving an IMR score of 1.39 and a PPL score of 184.8. This example shows the importance of selecting keywords that naturally fit the cue, rather than merely focusing on phonetic similarity and syllable count. Learners generally find cues that are more imageable and coherent to be more useful.



\noindent\textbf{Imageable but not Coherent}
For \textit{``exhaustive,''} the cue is useful because the keywords \textit{``horse''} and \textit{``stiff''} depict a vivid scene where a horse becomes tired after training. However, the phrase \textit{``horse turned stiff''} is not commonly used (usually referred as \textit{``stiff horse''}), leading to a higher PPL score. For \textit{``phenomena,''} even though the verbal cue is coherent and consists of simple keywords, the keywords do not contribute to depicting a concrete scene nor relate to the word's meaning. 
This example suggests that the keywords should be closely linked to the word's meaning, even if it slightly impacts the cue's coherence.

\noindent\textbf{Coherent but not Imageable}
For \textit{``retract,''} the cue is useful because it creatively describes a detective retracting his suspicion after reviewing a clean record, but the cue for \textit{``intimidate''} lacks creativity. Instead, the human-authored cue \textit{``An intimate date tends to intimidate her.''} is rated higher by learners across three aspects (imageability: $3.7$ vs. $3.0$, coherence: $3.8$ vs. $3.0$, usefulness: $3.9$ vs. $2.4$), evoking strong emotions to make it memorable. It is worth noting that the text-to-image model somewhat failed to describe the scenario in the cue for \textit{``retract,''} leading to a lower IMR score. This example suggests that certain properties of verbal cues, such as creativity or emotion, may overcome other problems such as a lack of imageability. 


\noindent\textbf{Less Imageable \& Coherent}
For \textit{``threadbare,''} the cue is useful because the context \textit{``worn to a thread''} successfully conveys the meaning of the target word. The cue for \textit{``virtuoso''} lacks an appropriate context, failing to indicate any connection to art. Regardless of the relevance of other generated cues containing art contexts like pianist or violin, they are not highly ranked due to their excessive length, illustrating a trade-off between suggestions ii) and iv) from the pilot study. Although learners think the verbal cue for \textit{``threadbare''} is both imageable and coherent, the metrics indicate otherwise, likely due to the abstract nature of \textit{``cherish,''} which is challenging to visualize and the ambiguity surrounding whether \textit{``bear''} refers to a teddy bear.
The discrepancies highlighted above between usefulness ratings and imageability/coherence of the verbal cue suggest that future work should focus on aligning automated metrics with human preferences, possibly using preference optimization~\cite{rafailov2024direct}, and improving the fidelity of verbal-visual conversion using techniques developed for abstract linguistic metaphors~\cite{chakrabarty2022}. 

\section{Related Work}
\label{sec:rw}

Imageability is defined as \textit{``the ease with which a word arouses sensory images''}~\cite{paivio1968concreteness}. To quantify this intangible aspect of language, psycholinguists and psychologists have compiled human imageability ratings databases like the MRC Psycholinguistic Database~\cite{wilson1988mrc} and Glasgow Norms~\cite{scott2019glasgow}. However, collecting this data through interviews is costly and time-consuming, making it difficult to scale psycholinguistic databases to the size of modern NLP corpora like the Corpus of Contemporary American English, which includes over 60,000 lemmas with frequency and parts of speech data~\cite{wu2023composition}.

Recent studies have addressed the challenge of scalability in assessing imageability by applying machine learning techniques to automate the collection of imageability ratings. These efforts include predicting imageability using supervised learning~\cite{ljubesic18predicting}, and employing image data mining to estimate word imageability by analyzing a various visual features~\cite{kastner2020estimating}. Moreover, advancements have extended these predictive models to the sentence level. For instance, researchers have developed methods to evaluate the visual descriptiveness of captions and introduced ways to calculate a sentence's imageability score based on the imageability scores of its constituent words~\cite{umemura2021tell}. Additionally, computational techniques have been proposed that utilize text-to-image models to generate images and measure sentence imageability by calculating the cosine similarity between the image and word embeddings~\cite{wu2023composition}.

\section{Conclusions and Future Work}
In this paper, we explored using large language models for the task of automated keyword mnemonics generation for vocabulary learning, via a novel overgenerate-and-rank method. Through both automated and human evaluation with both English teachers and learners, we found that our generated verbal cues are comparable to or better than human-authored ones, in terms of imageability, coherence, and usefulness. We also studied the intrinsic subjectivity among learners in our evaluation through qualitative feedback, which led to relatively lower inter-rater agreement. 

There are many avenues for future work. First, due to this intrinsic subjectivity, we need to develop methods to generate \emph{personalized} cues that each individual learner will find useful, by adapting to their personal language knowledge and cultural background. Second, we need to study automated \emph{visual} cue generation, in the form of images or even videos, and evaluate their quality separately from the verbal cues. Third, we plan to conduct an experiment with real language learners in classrooms, to evaluate whether our automatically generated verbal and even visual cues can indeed enhance vocabulary recall over a long period of time.

\clearpage

\section*{Limitations}


The study has a few limitations that should be considered. First, the scope of the study on Keyword Mnemonics is limited to English. Future studies could benefit from applying to second language acquisition. Second, a limited number of teachers and learners evaluating the verbal cues generated by the LLM might not be enough to gain a comprehensive understanding of their quality and usefulness. The opinions and insights of a larger pool of English experts would provide a broader range of perspectives and expertise, contributing to a more robust evaluation of the LLM-generated cues. Third, the study primarily assessed KM usefulness through teacher and learner preference ratings rather than practical application on long-term memory tests. While teacher and learner opinions provide valuable insights, a more robust evaluation would involve measuring the impact of KMs on actual long-term memory retention in language learners. Incorporating such practical assessments would accurately reflect the KMs' efficacy in real-world language learning scenarios. 

\section*{Ethics Statement}

Our human experiment was conducted with the approval of the Institutional Review Board (IRB). Prior to their participation, we provided teachers and learners with a consent form that thoroughly outlined the potential risks, benefits, time commitment, expected actions, and compensation. The compensation offered to teachers and learners adhered to the recommended amount for academic research studies, which is no less than the federal minimum wage of \$7.25 per hour.

\bibliography{acl_latex}

\appendix

\clearpage
\onecolumn
\section*{\Large Supplementary Material}
\label{sec:supp}

\section{Prompt}
\label{sec:prompt}

\begin{longtable}{|p{2.5cm}|p{12.5cm}|}
    \hline
    \centering Prompt & Game Name: Sounds Like A Word\newline

Game Objective:\newline
The objective of this linguistic game is to dissect a target word into syllables and find an array of common real words that, when pronounced, produce a similar sound to the target word. This game is not about spelling, but about phonetics.\newline

Game Rules:\newline
1. Each part of the target word should correspond with a standalone word, not a prefix or a suffix.\newline
2. The words selected should be suitable for those preparing for SAT, so players must choose words that are commonly used and understood at that level. This means that obscure, rare, or highly specific words should be avoided.\newline
3. The goal is to find words that sound like the target word when pronounced together, not necessarily to match the exact number of syllables in the target word. Therefore, the number of keywords might be less or more than the number of syllables in the target word. For instance, "duplicity" with four syllables could translate into "do please city" with three keywords.\newline
4. Players must correctly split each target word into syllables and locate common words that closely match these syllables in terms of sound. This game tests both phonetics knowledge and creativity.\newline
5. Players should avoid using slang, offensive, or sexually suggestive words as keywords. The chosen words should be appropriate and respectful.\newline

Here are the examples of how the input and output should look like:\newline

[Input]\newline
Target word: alleviate\newline
[Output]\newline
Keywords: a, leaf, he, ate\newline

[Input]\newline
Target word: belie\newline
[Output]\newline
Keywords: bee, lie\newline

[Input]\newline
Target word: duplicity\newline
[Output]\newline
Keywords: do, please, city\newline

[Input]\newline
Target word: appease\newline
[Output]
\\
\hline
\centering Response & 
Keywords: a, peas
\\
\hline
\caption{Prompts for generating syllabic keywords.}
\label{tab:key_prompt}
\end{longtable}
    
\begin{longtable}{|p{2.5cm}|p{12.5cm}|}
\hline
\centering Prompt & Game Description:\\
& \\
& In StoryWeave, players are given a target word and a set of keywords. The task for players is to craft an engaging story using these words cleverly. The ultimate challenge is to construct a narrative that not only incorporates the target word but also includes the keywords in the exact order presented. The beauty of the game lies in the players' use of imagination and language to unfold a thought-provoking plot and articulate characters.\\
& \\
& Upon completion of the narrative, players generate a summary of their narrative, with an emphasizing requirement: the keywords still must appear, but in the same sequence as provided in the original story. This twist lies in maintaining the faithfulness of the summary to the initial tale, allowing no room for distortion of the original plotline.\\
& \\
& In StoryWeave, creativity and logic flow in harmony, leading to captivating narratives that tie readers from a solitary word to an entire universe-shaped by strategic word placement and unrestricted imagination.\\
& \\
& Game Rules/Constraints:\\
& 1. Each player is handed a target word and a set of keywords.\\
& 2. The target word should be weaved into the story while the keywords are to appear exactly in the order given.\\
& 3. After crafting the narrative, a summary is to be produced. This summary will also utilize the keywords, keeping them in the identical order as featured in the initial story.\\
& 4. Players are strictly prohibited to rearrange the sequence of the keywords.\\
& 5. Both the full-length story and its summary should unequivocally reflect the original plotline, with no discrepancies allowed.\\
& \\
& The idea is to balance the ingenious use of vocabulary to construct a story and maintain absolute fidelity to the given keywords and their order. The essence of this game lies in the inventiveness of constructing an interesting narrative around the given `word constraints'.\\
& \\
& Here are the examples of how the input and output should look like:\\
& \\
& [Input]\\
& Target word: alleviate\\
& Meaning: relieve; make more bearable\\
& Keywords: a, leaf, he, ate\\
& [Output]\\
& Story: He found himself famished, left with nothing but a single leaf on his plate. With a sigh of resignation, he ate it, hoping to alleviate his gnawing hunger a bit.\\
& Summary: On his plate, there was a leaf he ate to alleviate his hunger.\\
& \\
& [Input]\\
& Target word: belie\\
& Meaning: contradict; give a false impression\\
& Keywords: bee, lie\\
& [Output]\\
& Story: In the sunlit meadow, Sunbeam, a deceitful bee, told an enchanting lie. His sweet tales belied his venomous sting, misleading creatures who fell prey to his deceptive harmony.\\
& Summary: The bee tells a lie that belies its nasty sting.\\
& \\
& [Input]\\
& Target word: appease\\
& Meaning: soothe; relieve\\
& Keywords: a, peas\\
& [Output]\\\hline
\centering Response & Story: The young boy was upset, he hated vegetables. His mother, understanding his distaste, offered him a plate of peas. She hoped this would appease his frustration and coax him into eating healthier.\\
& Summary: To appease the boy's dislike for vegetables, his mother offered him a plate of peas.\\
\hline
\caption{Prompts for generating verbal cues.}
\label{tab:verbal_prompt}
\end{longtable}

\begin{longtable}{|p{2.5cm}|p{12.5cm}|}
    \hline
    \centering Prompt & What would be the most likely word in to be in the [MASK]? List 5 candidates. \\
    & \\
    & To [MASK] the boy's dislike for vegetables, his mother offered him a plate of peas. \\
    \hline
    \centering Response & 1. Overcome \\
    & 2. Counteract \\
    & 3. Combat \\
    & 4. Challenge \\
    & 5. Confront \\
    \hline
    \caption{Prompts for generating candidates for context completeness ranking ($R_\text{cont}$).}
    \label{tab:mask_prompt}
\end{longtable}

\section{Methodology}
\label{sec:supp_methodology}
This section provides a detailed description of the criteria employed for the ranking.

\subsection{Keyword Ranking}
For the assessment of imageability ($R_{\text{img}}$) and semantic similarity ranking ($R_{\text{sem}}$), lemmatization was applied to all keywords and target words. Imageability scoring involved the utilization of two psycholinguistic rating datasets: Glasgow Norms~\cite{scott2019glasgow} and the dataset presented in~\citet{ljubesic18-predicting}. We prioritize Glasgow Norms ratings and re-scaled ~\citet{ljubesic18-predicting} to $7$-point scale for consistency. During the computation of the average imageability score, stopwords were excluded. Words not present in the dataset were assigned a score of 1, representing the minimum value on the scale.

\subsection{Verbal Cue Ranking}
For the assessment of context completeness and AoA ranking, lemmatization was applied to all keywords and target words. AoA scoring involved the utilization of the rating dataset: \citet{kuperman2012age}. During the computation of summed AoA score, stopwords were excluded. Words not present in the dataset were disregarded.

\clearpage
\section{Verbal Cues}
\label{sec:verbal_cues}
Human-authored cues are from \citet{Geer2018picturethese}, a work copyrighted by Barron's Educational Series, Inc. LLM-generated cues and human-authored cues are only used for human evaluation. 

\begin{longtable}{|c|p{6.5cm}|p{6.5cm}|}
\hline
Word  & \multicolumn{1}{c|}{LLM-generated} & \multicolumn{1}{c|}{Human-authored} \\ \hline
abstemious & \textit{Ab}'s diet of only the \textit{stem} of vegetables taught \textit{me} and \textit{us} to be more \textbf{abstemious}. & \textit{Ab}'s \textit{steamy} with \textit{us} when he tells us to be \textbf{abstemious}. \\\hline
accolade & The \textit{echo} of the \textit{lad}'s music earned him an \textbf{accolade}. & \textit{A} \textit{coal} \textit{aide} receives an \textbf{accolade}. \\\hline
aesthetic & \textit{As} \textit{the} clock \textit{ticked}, he was captivated by the \textbf{aesthetic} appeal of the painting. & Oh beautiful \textit{S}! \textit{That} \textit{tick} tickles the \textbf{aesthetic} sense! \\\hline
appease & He cooks \textit{a} dish with \textit{peas} to \textbf{appease} her anger. & Tom was \textbf{appeasing} a pot \textit{o'} \textit{peas}. \\\hline
archaic & The twins wonder ``\textit{are} these \textit{cake} instructions?'' from an \textbf{archaic} recipe book. & Sick kangaroos ride on \textbf{archaic} \textit{Ark} \textit{K} - ``\textit{Ick}!'' is all they can say. \\\hline
artisan & In the \textit{art}-loving town, the \textbf{artisan} sips his \textit{tea} under the \textit{sun}. & ``\textit{Art} \textit{is} \textit{sin},'' says the Puritan to the \textbf{artisan}. \\\hline
ascendancy & The \textit{ass} and \textit{hen} \textit{dance} by the \textit{sea} to determine their \textbf{ascendancy}. & The Egyptians are doing the \textit{Ascend} \textit{Dance} to \textbf{ascendancy}.  \\\hline
authoritarian & The citizens of \textit{Ought} face the choice to submit \textit{or} resist the \textbf{authoritarian} regime, causing their freedom to \textit{tear} like \textit{rain}. & \textbf{Authoritarian} \textit{Arthur} \textit{Tarian} tells Ian to tear up other authors' works. \\\hline
beguile & A \textit{bee} and a \textit{guy} \textbf{beguile} people in each supermarket \textit{aisle} they walk down together. & Be Guy well, or \textit{be} \textit{Guy} \textit{ill}, he must \textbf{beguile} the nurse from Brazil. \\\hline
censure & John's \textit{sin} made his mother \textit{sure} to \textbf{censure} him. & A \textit{cent} with a \textit{sore} \textbf{censured} at the Century Bank. \\\hline
compliance & Using a \textit{comb} to \textit{ply} his hair, Billy observes \textit{ants'} \textbf{compliance} to their queen's orders. & ``\textit{Come}, \textit{ply} \textit{ants} with honey to gain their \textbf{compliance} with our plan.'' \\\hline
concede & The \textit{con} man was outsmarted by a humble \textit{seed} seller, leading him to \textbf{concede} defeat. & ``I \textbf{concede} that it's \textit{cons'} \textit{seed}, but I need feed!'' \\\hline
conciliatory & A \textit{con} man helps a wounded \textit{seal} \textit{he} found, \textit{tearing} his clothes for a bandage in a \textbf{conciliatory} act, leading to his \textit{re}formation. & The \textit{con's} \textit{silly} \textit{story} - a \textbf{conciliatory} gesture. \\\hline
contend & The city's \textit{con} artist \textit{tends} to \textbf{contend} that he's misunderstood. & He \textbf{contends} that a \textit{con} \textit{tends} to return. \\\hline
conventional & \textit{Con} finds a \textit{van}, and against his \textbf{conventional} nature to \textit{shun} \textit{all} rules, reports it. & At a \textit{convention}, \textit{all} must wear \textbf{conventional} suits. \\\hline
conviction* & \textit{Con} artist \textit{Victor}'s \textbf{conviction} was proven by the necklace found against his \textit{shin}. & ``It is my \textbf{conviction} that every \textit{convict} should \textit{shun} guns.'' \\\hline
cordial & In the city's \textit{core}, the \textbf{cordial} shopkeeper makes fair \textit{deals}. & ``I've cut the \textit{cord}, \textit{Jill}!'' (Relations are no longer \textbf{cordial}.) \\\hline
cryptic & In the \textit{crypt}, a mysterious `\textit{tick}' sound was a \textbf{cryptic} secret. & A \textbf{cryptic} \textit{crypt} \textit{tick}. \\\hline
degradation & The \textit{degrade} endured by the \textit{Asian} immigrant led to his personal \textbf{degradation}. & \textbf{Degradation} from a ``\textit{D}'' \textit{grade} \textit{date} - \textit{shun} it! \\\hline
depravity & The \textit{deep}, \textit{rave}-filled \textit{city} hides a world of \textbf{depravity}. & \textit{Deep} in the \textit{rabbit} warren, \textit{he} tells them a tale of \textbf{depravity}. \\\hline
deprecate & Luna looks at the \textit{deep} sea \textit{wreck} with \textit{hate}, \textbf{deprecating} man's recklessness. & Near \textit{Deep-Wreck} \textit{8}, they \textbf{deprecate} the ``Catch of the Day.'' \\\hline
disputatious & At ``\textit{Dis} Church'', the \textit{pew} is filled with \textbf{disputatious} locals, until Mrs. \textit{Tat} attempts to \textit{shush} them. & \textit{Dispute} \textit{8}. \textit{Just} another dispute between \textbf{disputatious} dates. \\\hline
divergent* & The \textit{ant}, on the \textit{verge} of \textit{dying}, faced \textbf{divergent} paths. & \textit{Di} on the \textit{verge} of going with a \textit{gent} on a \textbf{divergent} path. \\\hline
egotism & Mr. \textit{Ego}, over his \textit{tea}, tallies up the \textit{sum} of his achievements, revealing his \textbf{egotism}. & The main concern of \textit{E} \textit{goat} \textit{is} \textit{himself}. What \textbf{egotism}! \\\hline
emulate & The usually punctual \textit{emu} was \textit{late}, yet other birds still sought to \textbf{emulate} its habits. & ``\textit{Em}, \textit{you}'re \textit{late}! Must you \textbf{emulate} girls who make their dates wait?'' \\\hline
enmity & The \textit{hen} displays her \textbf{enmity} towards the farmer's golf \textit{mitt} at the \textit{tee}. & The \textit{N} \textit{mitt} \textit{he} wears causes \textbf{enmity}. \\\hline
exhaustive & An \textit{ex}-champion \textit{horse} turned \textit{stiff} from his \textbf{exhaustive} training. & ``Our \textbf{exhaustive} battery of tests does not \textit{exhaust} \textit{Steve}.'' \\\hline
feasible & With the rise in \textit{fees}, John was still \textit{able} to pay, making it a \textbf{feasible} solution for him. & It's not \textbf{feasible} to pay \textit{fees} to \textit{a} \textit{bull}. \\\hline
flagrant & The king's \textbf{flagrant} \textit{flaw} was to \textit{grant} favors indiscriminately, causing outrage. & \textbf{Flagrant} hostility at a \textit{flag} \textit{rant}. \\\hline
gullible & The \textit{gull}, by moving his \textit{lip}, convinces the \textbf{gullible} \textit{bull} of his stories. & ``If you believe in the \textit{gull-a-bull}, you must be \textbf{gullible}!'' \\\hline
ignominy & The man with an \textit{egg} has \textit{no} \textit{money}, embodying his \textbf{ignominy}. & \textit{A} \textit{gnome}, \textit{Minnie}, suffers no \textbf{ignominy}. \\\hline
illusory & \textit{Ill} and fearing he will \textit{lose} everything, John says \textit{sorry} in his \textbf{illusory} world. & \textit{A} \textit{Lew}, \textit{sorry} for having followed an \textbf{illusory} dream. \\\hline
implement & An \textit{imp}, after a deep \textit{lament}, decides to \textbf{implement} changes to his behavior. & They \textbf{implement} a curfew to an \textit{imp} \textit{lament}. \\\hline
inclusive & \textit{In} the search for a \textit{clue}, the detective's \textbf{inclusive} method acted like a \textit{sieve}, filtering through all the evidence. & Look \textit{in} \textit{clues} \textit{if} you seek \textbf{inclusive} evidence. \\\hline
inconsequential & The artist finds \textit{ink} \textit{on} his \textit{sequence} of strokes but remains \textit{chill}, deeming it \textbf{inconsequential}. & ``It's \textbf{inconsequential} how we go, but \textit{in} \textit{con} \textit{sequence} \textit{shall} we go if you insist!'' \\\hline
incorrigible & The artist used \textit{ink} for his drawing of an \textit{or}-like \textit{ridge} and a \textit{bull}, but the bull was an \textbf{incorrigible} mistake. & Trying to \textit{encourage} \textit{a} \textit{bull} to cease his \textbf{incorrigible} ways. \\\hline
indifferent & He was \textbf{indifferent}, even \textit{in} a \textit{different} situation. & People \textit{in} \textit{different} lands being \textbf{indifferent} to each other. \\\hline
ingenious & Dr. Jensen found an \textbf{ingenious} solution \textit{in} the \textit{gene}, saying `\textit{yes}' to his eureka moment. & \textit{In} \textit{genius} we find \textbf{ingenious} ideas. \\\hline
intimidate & ``\textit{In} a \textit{tea} gathering, a \textit{mate} walked in to \textbf{intimidate} everyone.'' & An \textit{intimate} \textit{date} tends to \textbf{intimidate} her. \\\hline
intrepid & \textit{In} his \textit{trip}, the \textbf{intrepid} explorer fearlessly faced what \textit{hid} in the shadows. & Even the most \textbf{intrepid} explorer should, \textit{in} his \textit{trip}, \textit{heed} warnings. \\\hline
oblivion & Joe wished, ``\textit{Oh}, \textit{live} \textit{beyond} this hardship,'' but his dreams sank into \textbf{oblivion}. & \textit{Ob} \textit{lived} and \textit{eon} before Oblantis fell into \textbf{oblivion}. \\\hline
opportunist & Always looking \textit{up} for a \textit{port} of advantage, \textit{you} would see an \textbf{opportunist} building his \textit{nest} on others' misfortunes. & At \textit{a} \textit{port} near \textit{Tunis}, an \textbf{opportunist} waits. \\\hline
opulence & An overwhelmed man exclaims "\textit{Oh}!" as he sits at a \textit{pew}, observing the cathedral's \textbf{opulence} through his \textit{lens}. & ``An \textit{opal} \textit{lance}? Such \textbf{opulence} in this palace!'' \\\hline
peripheral & A \textit{pair} teased with `what \textit{if} \textit{her} \textit{doll} disappears?', hiding it in \textbf{peripheral} areas, but her attention captured it all. & ``There's a \textit{pear} \textit{for} \textit{all} on the \textbf{peripheral} pear trees!'' \\\hline
phenomena & Sarah, studying \textbf{phenomena}, was on the \textit{phone} when she had to say, ``\textit{No}, \textit{Ma}''. & ``And now its a \textit{fin} \textit{omen} - \textit{ahhh}! - rare \textbf{phenomena}'' \\\hline
polemical & The \textbf{polemical} John, standing like a \textit{pole}, accuses \textit{me} of having ideas as dark as \textit{coal}. & A \textbf{polemical} \textit{polar} \textit{Mick} \textit{call}. \\\hline
quiescence & The \textit{key} to peace is in the \textit{essence} of \textbf{quiescence}. & \textit{Qwee} \textit{Essence} brings \textbf{quiescence}. \\\hline
rant & She \textit{ran} out of patience waiting for her \textit{tea} and went on a \textbf{rant}. & ``By \textit{Ra}, that \textit{ant} can \textbf{rant}!'' \\\hline
ratify & The \textit{rat} agrees to \textbf{ratify} a \textit{tie}-up treaty for a \textit{fee}. & ``I'll be a \textit{rat} \textit{if} \textit{I} \textbf{ratify} this treaty.'' \\\hline
recount & Mathematician \textit{Ree} had to \textit{count} and then \textbf{recount} to secure her win. & Recounting how \textit{Rick} was picked to \textit{count} the votes in the \textbf{recount}. \\\hline
rectify & Wearing his \textit{red} \textit{tie}, he decided to \textit{fie} and \textbf{rectify} his mistake. & ``I'll be \textit{wrecked} \textit{if} \textit{I} don't \textbf{rectify} my neighbor's behavior.'' \\\hline
rescind & The courier had the letter on his \textit{wrist} to \textit{send}, but the order was \textbf{rescinded}. & \textit{Reese} \textit{sent} his team a memo to \textbf{rescind} his previous one. \\\hline
retract & After \textit{reading} the clean \textit{track} record of the suspect, the detective had to \textbf{retract} his suspicion. & \textit{Rhet} \textit{tracked} through the woods to \textbf{retract} his words. \\\hline
rigor & On the \textit{ridge}, John says ``\textit{Er}...'' but the \textbf{rigor} of his training encourages him to continue. & The \textit{rig} is checked by captain \textit{Gore} with \textbf{rigor}. \\\hline
stoic & A fire starts from the \textit{stove}, to which the \textbf{stoic} chef merely reacts with `\textit{ick}'. & \textbf{Stoic} customers tuck at the \textit{Stowe} \textit{Wick} in Stowe, Vermont. \\\hline
surreptitious & `\textit{Sir}', a `\textit{rep}' manager, needs to hide a scandalous \textit{titbit} with a \textbf{surreptitious} `\textit{shush}'. & \textit{Sir} \textit{Repetitious} in a \textbf{surreptitious} operation. \\\hline
tantamount & The \textit{ant} finds a \textit{tea} \textit{mound} \textbf{tantamount} to a valuable treasure. & Using \textit{Tant} as \textit{a} \textit{mount} to ascend El Pico in Peru is \textbf{tantamount} to saying alpacas' rights are few! \\\hline
threadbare & Worn to a \textit{thread}, the \textit{bear} became \textbf{threadbare} but still cherished. & \textit{Thread} \textit{Bear} takes orders from the \textbf{threadbare} customers. \\\hline
unwarranted & An unexpected `\textit{un}' \textit{war} \textit{ran} through \textit{Ted}'s land, leading to his \textbf{unwarranted} accusation. & ``\textit{A} \textit{warrant}?'' \textit{Ted} asked. ``That's \textbf{unwarranted}.'' \\\hline
virtuoso & \textit{Veer}'s talent was \textit{too} remarkable, \textit{so} he became a \textbf{virtuoso}. & This \textbf{virtuoso} has a \textit{virtue} \textit{oh} \textit{so} rare - he spreads cheer far and near. \\\hline
\caption{Verbal cues used for human evaluation. Keywords are represented in \textit{italic}, while a target word is in \textbf{bold}. * indicates an anomaly in verbal cue generation using LLM.}
\label{table:my_longtable} 
\end{longtable}

In the case of the target word ``conviction,'' only one set of keywords (``con,'' ``vict,'' ``shun'') were generated. In the case of the target word ``divergent,'' the model failed to arrange the keywords correctly, which should have shown in the order of (``die,'' ``verge,'' ``ant'').

\clearpage
\section{Human Evaluation}
\label{sec:sp_human_eval}
\subsection{Web Interface}
\begin{minipage}{\textwidth}
    \strut\newline
    \centering        
    \includegraphics[width=\linewidth]{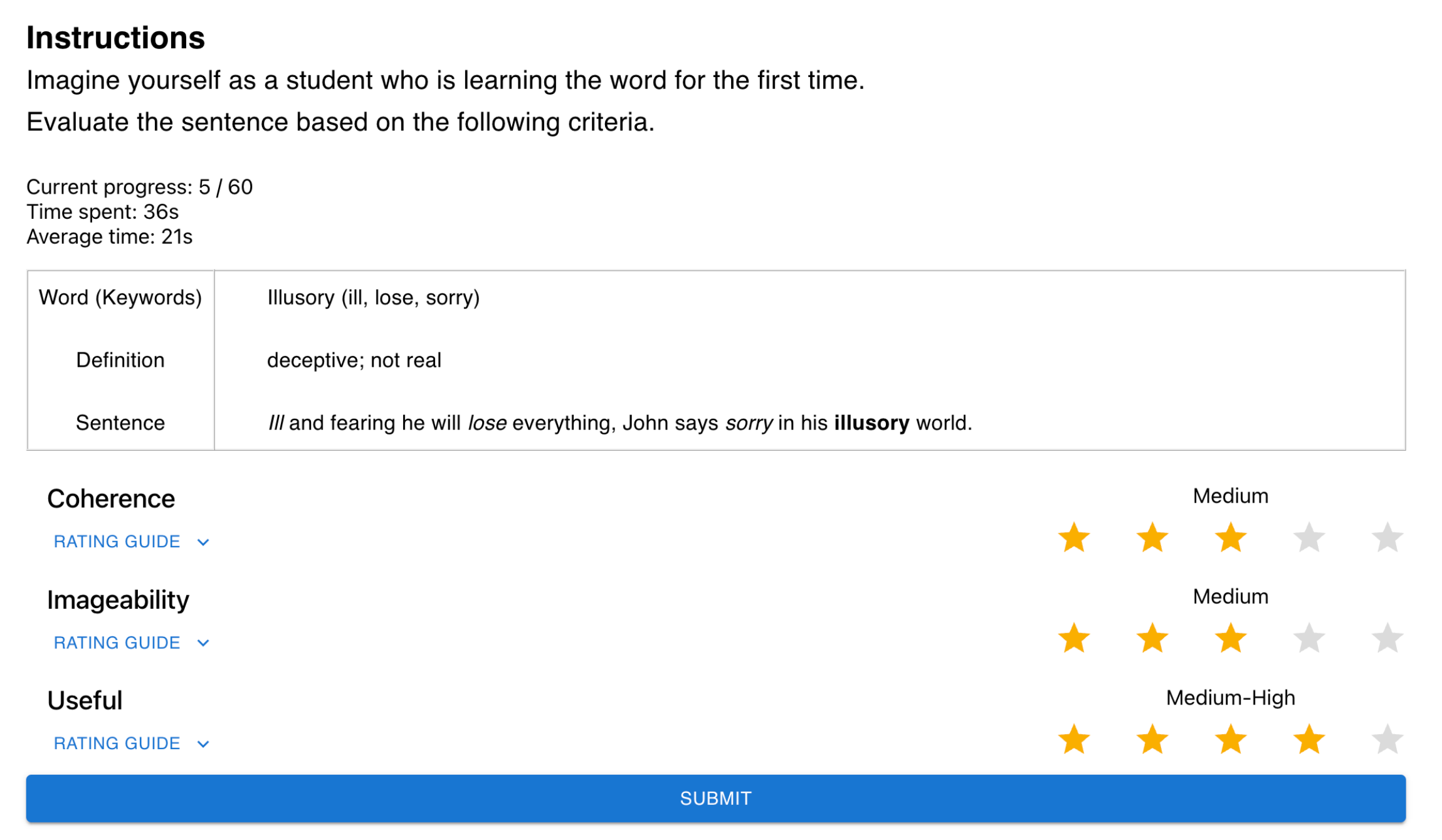}
    \captionof{figure}{Web application interface for human evaluation.}
    \label{fig:user_interface}
\end{minipage}

\subsection{Criteria}
Guidelines for rating 5-point Likert Scale on imageability, coherence, and usefulness.  

\begin{minipage}{\textwidth}
\centering
\strut\newline
\scalebox{0.85}{
\begin{tabular}{>{\raggedright}m{2cm} >{\raggedright\arraybackslash}m{10cm}}
\toprule
Scale      & Explanation \\ \midrule
High (5)   & The sentence evokes a vivid and detailed mental image, making it easy to visualize the scene or situation described in the sentence. \\
Medium (3) & The sentence evokes a reasonable level of imagery: minor inconsistencies may exist in the description, but a mental image can still be formed. \\
Low (1)    & The sentence lacks substantial imagery, making it challenging to form any meaningful mental image. \\ \bottomrule
\end{tabular}
}
\captionof{table}{Instructions for rating \textbf{imageability} of the verbal cues.}
\label{tab:scoring_imageability}
\end{minipage}

\begin{minipage}{\textwidth}
\centering
\strut\newline
\scalebox{0.85}{
\begin{tabular}{>{\raggedright}m{2cm} >{\raggedright\arraybackslash}m{10cm}}
\toprule
Scale      & Explanation                                                                                                            \\ \midrule
High (5)   & The sentence is highly coherent: the meaning is clear, and the wording is natural.                                      \\
Medium (3) & The sentence is moderately coherent: minor issues affect clarity in understanding the meaning.                         \\
Low (1)    & The sentence lacks coherence: it’s difficult to understand or read because it is illogical or grammatically incorrect. \\ \bottomrule
\end{tabular}
}
    \captionof{table}{Instructions for rating \textbf{coherence} of the verbal cues.}
    \label{tab:scoring_coherence}
\end{minipage}

\begin{minipage}{\textwidth}
\centering
\strut\newline
\scalebox{0.85}{
\begin{tabular}{>{\raggedright}m{2cm} >{\raggedright\arraybackslash}m{10cm}}
\toprule
Scale      & Explanation                                                                                                            \\ \midrule
High (5)   & The sentence given is a useful tool for memorizing the vocabulary word. It will help me remember the meaning of the word, or I imagine it would be helpful to others.                                      \\
Medium (3) & The sentence provided has issues that affect how useful I find it, but with some minor modifications, it could be useful.                         \\
Low (1)    & This sentence is not useful at all. \\ \bottomrule
\end{tabular}
}
    \captionof{table}{Instructions for rating \textbf{usefulness} of the verbal cues.}
    \label{tab:scoring_usefulness}
\end{minipage}

\clearpage

\section{Ablation Study}
\label{sec:sp_ablation}






\subsection{Llama3 Fine-tuning}
The book ``Picture These SAT Words!'' includes a total of 314 words. For our study, we allocated 60 of these words to a test set for human evaluation. On the remaining words, we perform an 80-20 train-validation set split to fine-tune Llama3 for the KM generation task. We excluded the words ``alleviate'' and ``belie'' from this training set as they were used as in-context examples. We leverage an early-stopping approach where we evaluate the model on the validation set after each epoch and utilize the model weights with the highest validation performance at test time.

Specifically, we utilize the \texttt{meta-llama/Meta-Llama-3-8B} as the base model, paired with the Adam optimizer, set to its default parameters (\texttt{b1}=$0.9$, \texttt{b2}=$0.999$, \texttt{eps}=1e-6). We utilize an learning rate of 2e-5 and batch size of $2$ for $5$ epochs. Instead of full fine-tuning we use, LoRA adaptors for the modules \texttt{q\_proj}, \texttt{k\_proj}, \texttt{v\_proj}, and \texttt{o\_proj}, with each adaptation characterized by a rank $r$ of 16, alpha of $16$, and dropout rate of $0.1$.

\subsubsection{Prompt}
\begin{longtable}{|p{2.5cm}|p{12.5cm}|}
    \hline
    \centering Prompt & Your task is to create memorable keywords and a sentence that helps memorize a specific target word. The keywords should consist of words that resemble the phonetic sounds of the target word's syllables. The sentence should incorporate the provided keywords and the target word. Please adhere to the following rules:\newline

1. Keep the original keyword order in the sentence.\newline
2. Provide clear context for the target word.\newline
3. Use words at the same (complexity) level as or lower than the target word.\newline
4. Keep the cue short; long ones are not helpful.\newline

[Input]\newline
Target word: alleviate\newline
Meaning: relieve; make more bearable\newline
Keywords: a, leaf, he, ate\newline
[Output]\newline
Sentence: On his plate, there was a leaf he ate to alleviate his hunger.\newline

[Input]\newline
Target word: belie\newline
Meaning: contradict; give a false impression\newline
Keywords: bee, lie\newline
[Output]\newline
Sentence: The bee tells a lie that belies its nasty sting. \newline

[Input]\newline
Target word: appease\newline
Meaning: soothe; relieve\newline
Keywords: a, peas\newline
[Output]\\
    \hline
    \centering Response & Sentence: To calm the toddler, a peas dish was made to appease her cries. \\
    \hline
    \caption{Prompt for ablation $\text{Llama3}_{\text{FT}}$.}
    \label{tab:llama_ft_prompt}
\end{longtable}



\subsubsection{Llama3 Sampling Configuration}

For both fine-tuning and zero-shot, we perform nucleus sampling of Llama3. We set a \texttt{temperature} of $1.0$, ensuring a balanced approach to novelty and feasibility in outputs. We set \texttt{top\_p} to $0.95$, which allows the model to consider a range of token possibilities, enhancing creativity without straying too far from plausible completions. Furthermore, \texttt{top\_k} is restricted to $50$, focusing the model’s choices to the top $50$ most probable next tokens, which helps in maintaining coherence and relevance in the generated text.

\section{Software Package}

In our study, we employed various software packages. We used \texttt{spearmanr} and \texttt{wilcoxon} from \texttt{scipy.stats} for calculating Spearman's rank correlation coefficient and performing the Wilcoxon signed-rank test, respectively. For word lemmatization, the \texttt{nltk} package was utilized. Additionally, we used \texttt{lmppl}, specifically \texttt{meta-llama/Meta-Llama-3-8B}, for computing perplexity.

\clearpage

\end{document}